\documentclass[runningheads]{llncs}

\usepackage[T1]{fontenc}
\usepackage{graphicx}
\usepackage{booktabs}
\usepackage{amsmath}
\usepackage{xcolor}
\usepackage{multirow}
\usepackage{pgfplots}
\pgfplotsset{compat=1.18}
\usepgfplotslibrary{groupplots}

\usepackage{subcaption}

\usepackage{xcolor}
\newcommand\crule[3][black]{\textcolor{#1}{\rule{#2}{#3}}}

\usepackage[hidelinks]{hyperref}

\definecolor{cBaseline}{HTML}{6b7280}
\definecolor{cEnsemble}{HTML}{15803d}
\definecolor{cDropout}{HTML}{b91c1c}
\definecolor{cDropConnect}{HTML}{0369a1}
\definecolor{cFlipout}{HTML}{7e22ce}
\definecolor{cDUQ}{HTML}{ca8a04}

\title{Uncertainty Identifies Difficult Samples Across Methods: A Multi-Task Study on a Heterogeneous Skin Lesion Dataset}

\titlerunning{UQ Identifies Difficult Samples in Skin Lesion on a Heterogeneous Dataset}

\author{Leon Koole \and Jiapan Guo \and Matias Valdenegro-Toro}
\institute{Bernoulli Institute, University of Groningen, Netherlands}

\begin{document}

\maketitle

\begin{abstract}

Skin lesion classifiers can be confidently wrong on the cases that matter most, so knowing when a prediction should not be trusted is clinically as useful as the prediction. We study uncertainty quantification on a dataset pooled from many ISIC sources, with a shared backbone and two jointly learned heads: a binary malignant versus non-malignant head and a five-class diagnostic head. Five UQ methods (MC Dropout, DropConnect, Flipout, Deep Ensembles, DUQ) are compared on accuracy, calibration, uncertainty decomposition, and risk-coverage. Difficulty is largely method-agnostic: even methods with narrow entropy distributions rank the same samples as hard (per-sample entropy correlations of $0.54$ to $0.91$). The choice of method matters more for calibration and uncertainty decomposition, where Deep Ensembles is the clear winner, than for finding difficult cases. The ranking is also good enough that deferring the most uncertain cases removes a disproportionate share of errors, supporting uncertainty-based selective referral, evaluated here in-distribution only.

\end{abstract}

\section{Introduction}
\label{sec:intro}

Skin cancer is the most common cancer worldwide, and the hard part is telling the few malignant lesions, especially melanoma, from the many benign moles that look like them. Localized melanoma has a five-year survival above 99\%, dropping to a third once it spreads~\cite{acs_melanoma_survival}. Dermoscopy beats naked-eye inspection, yet benign and malignant lesions can look nearly identical, and histopathology, the reference standard, is invasive and used only for already-suspicious cases~\cite{dinnesDermoscopyVisualInspection2018}. Since Esteva et al.~\cite{esteva2017dermatologist}, deep learning has matched dermatologist-level accuracy on curated benchmarks, but malignant lesions are rare, so class imbalance pushes a classifier toward the majority~\cite{cassidyAnalysisISICImage2022}, and models trained mostly on light-skinned populations do worse on darker skin~\cite{daneshjouDisparitiesDermatologyAI2022,grohEvaluatingDeepNeural2021}. Aggregate accuracy says nothing about which single prediction to trust, and softmax confidence is not reliable~\cite{galUncertaintyDeepLearning}.

Uncertainty quantification (UQ) lets a classifier flag and abstain on its least confident cases and defer them to a clinician~\cite{geifmanSelectiveClassificationDeep2017}. Several families produce them: Bayesian approximations perturb the network across repeated forward passes, deep ensembles~\cite{lakshminarayananSimpleScalablePredictive} train independent networks, and single-pass methods such as DUQ~\cite{van2020uncertainty} read it from one pass, their trade-offs widely surveyed~\cite{gawlikowskiSurveyUncertaintyDeep2023,abdarReviewUncertaintyQuantification2021}.

In dermatology, UQ is mostly studied with one to three methods from the same family, on a single curated benchmark \cite{van2019quantifying}. Mobiny et al.~\cite{mobinyRiskAwareMachineLearning2019} built a Bayesian referral workflow, and similarly Van Molle et al. \cite{van2019quantifying} uses Bayesian Neural Networks. Combalia et al.~\cite{combaliaUncertaintyEstimationDeep2020} showed Monte Carlo sampling flags difficult and out-of-distribution cases on ISIC. Abdar et al.~\cite{abdarUncertaintyQuantificationSkin2021} compared dropout and ensemble variants for referral. These confirm that uncertainty can separate reliable from unreliable predictions, but rarely compare families on one model, and test on clean, single-source data. How UQ methods behave on pooled data, and whether they agree on which cases are hard, is much less clear.

We study this on a deliberately heterogeneous setting: a large test set pooled from many sources in the ISIC Archive~\cite{cassidyAnalysisISICImage2022}, spanning dermoscopic, clinical, and total-body-photography images at roughly a 70/30 benign/malignant split. We use this pooled dataset because it introduces realistic variation, clinical (case severity, labelling workup) and technical (imaging protocol, background), that an uncertainty estimate should reflect. We train five UQ methods plus a softmax baseline on a shared EfficientNet-B3 backbone with two heads: binary malignant-vs-non-malignant and five-class over consolidated categories.

\begin{figure}[t]
\centering
\newcommand{\tcell}[2]{%
  \begin{minipage}[t]{0.152\linewidth}\centering
  \includegraphics[width=\linewidth]{#1}\\[2pt]
  {\scriptsize #2}\end{minipage}}
\tcell{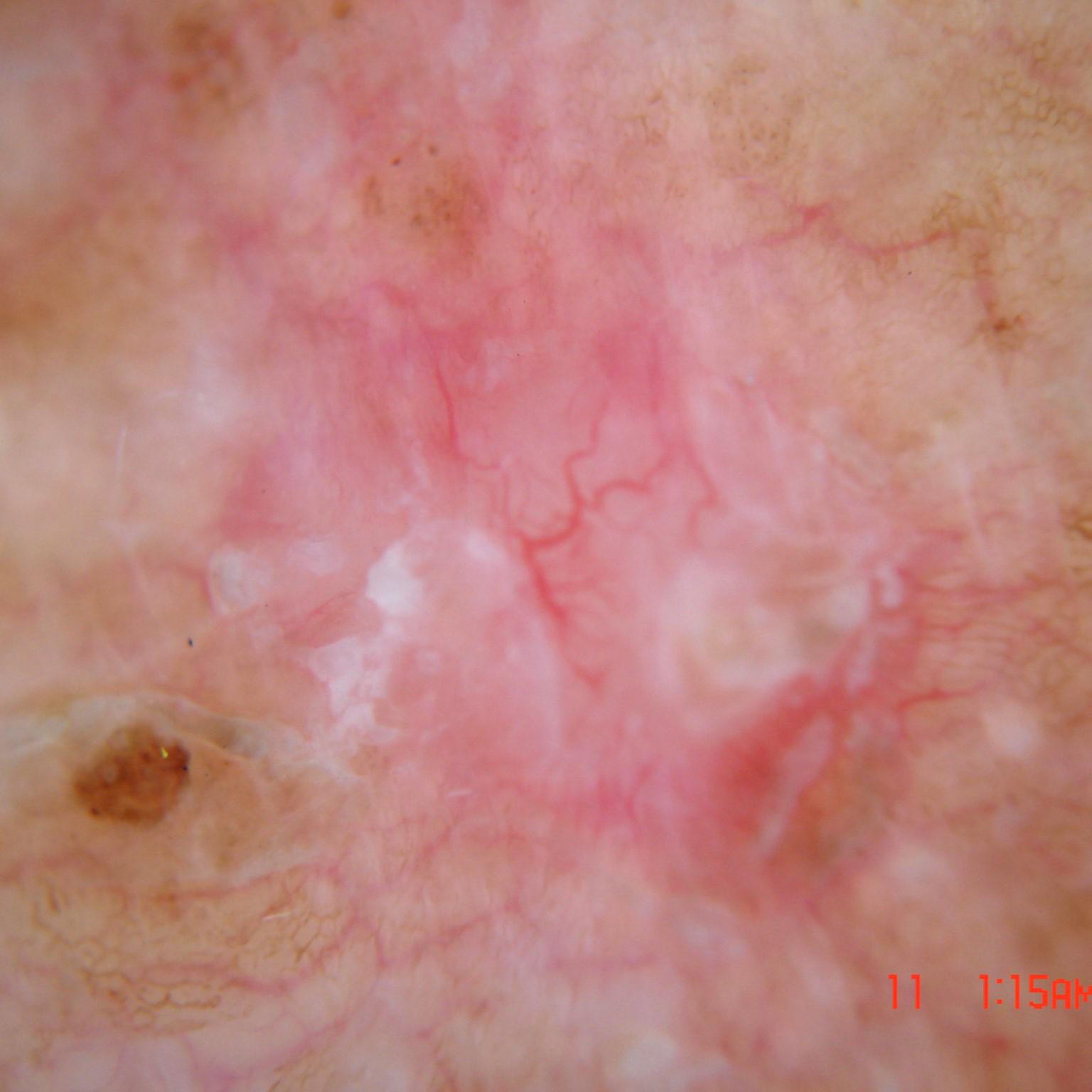}{Tier 0\\(all correct)}\hfill
\tcell{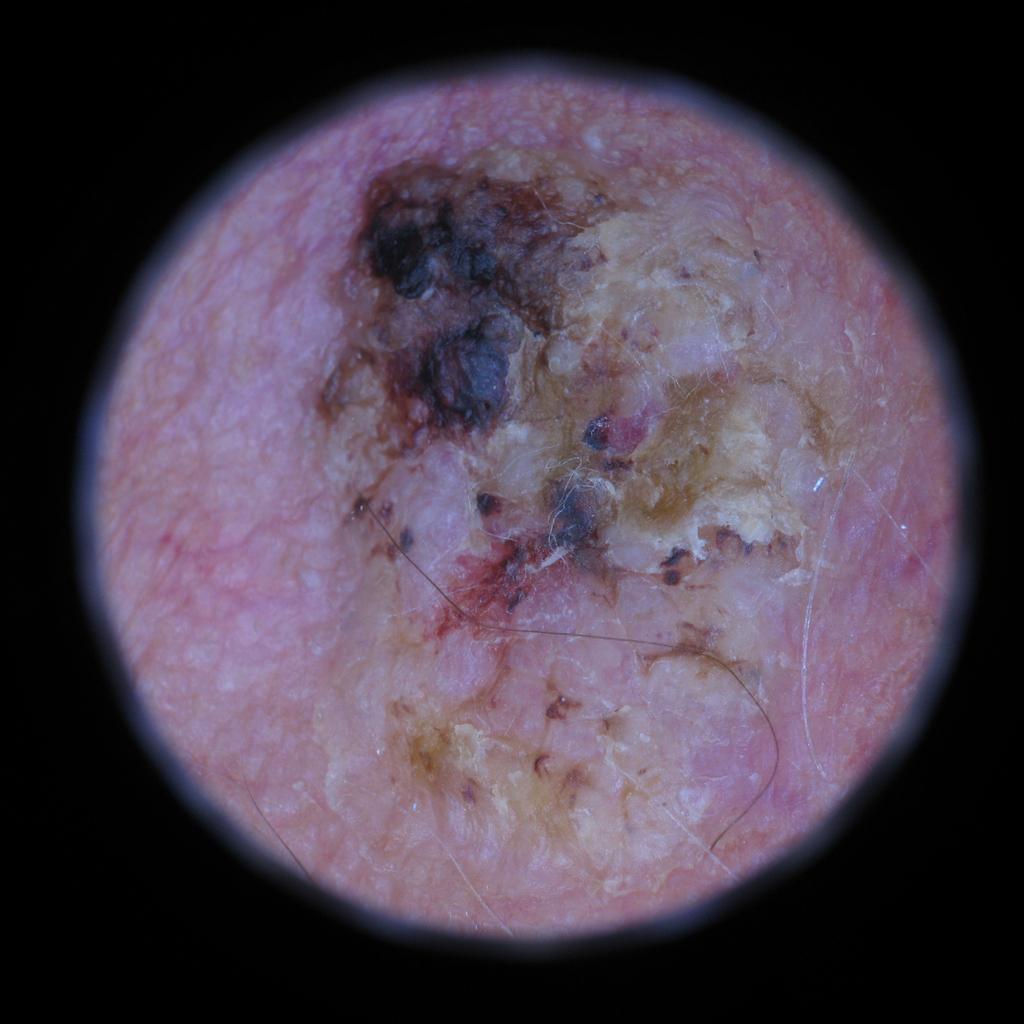}{Tier 1}\hfill
\tcell{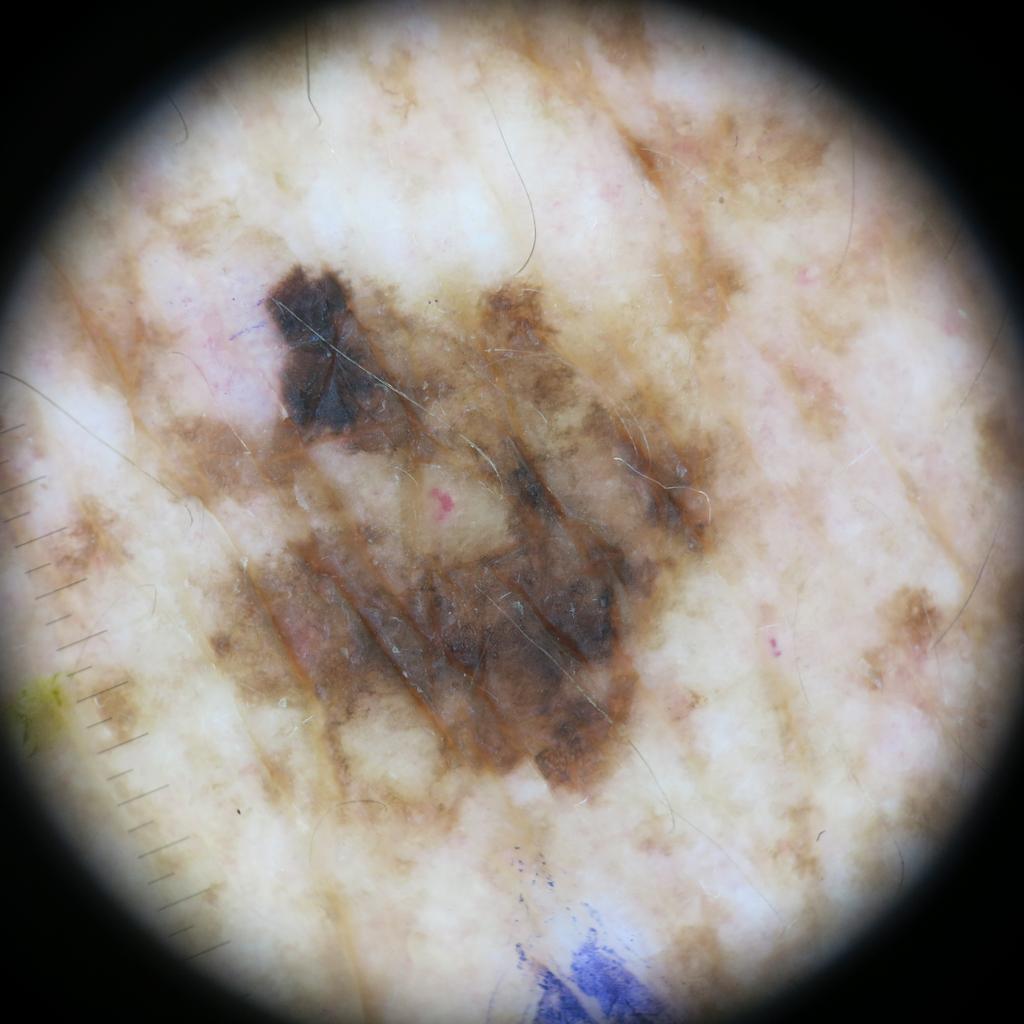}{Tier 2}\hfill
\tcell{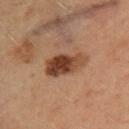}{Tier 3}\hfill
\tcell{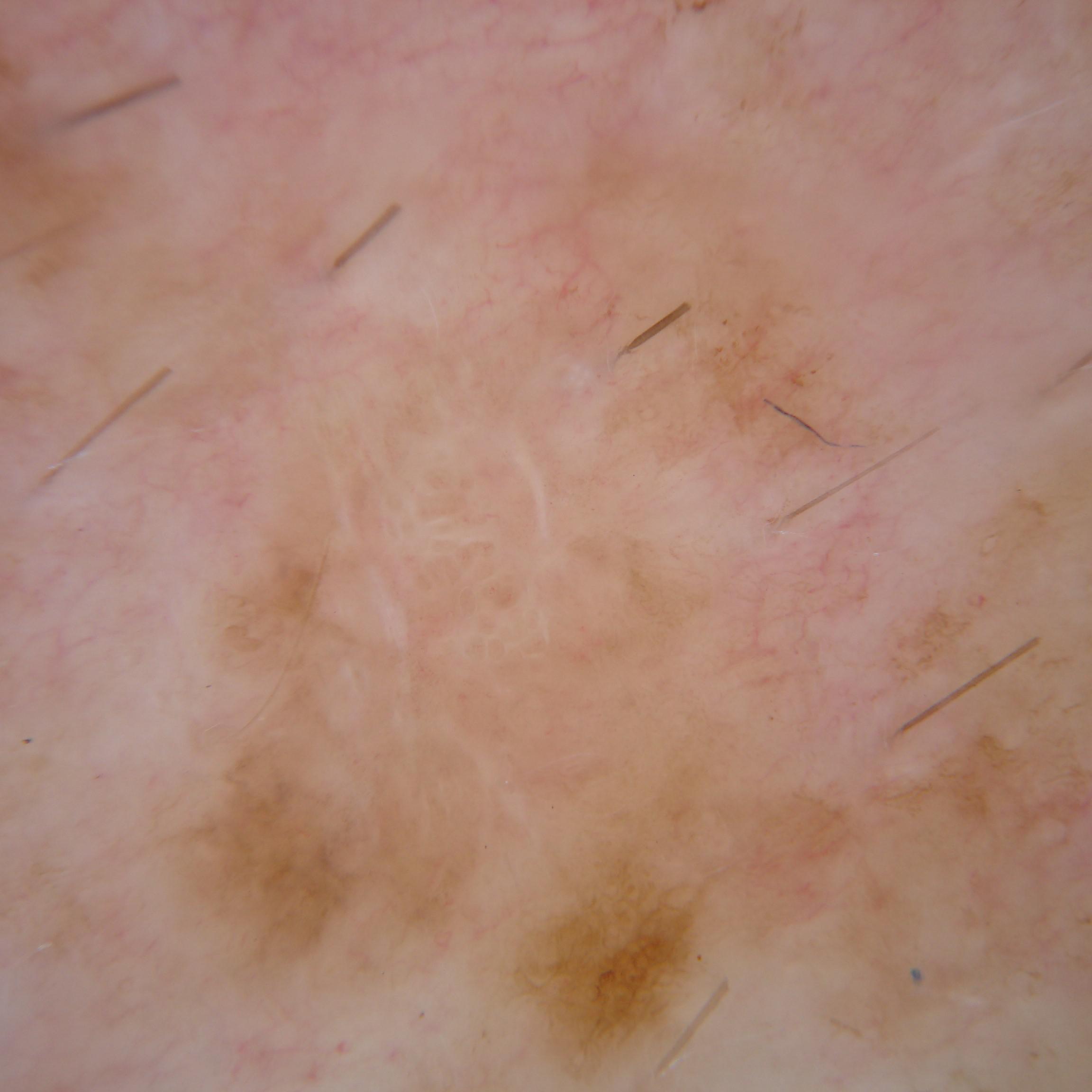}{Tier 4}\hfill
\tcell{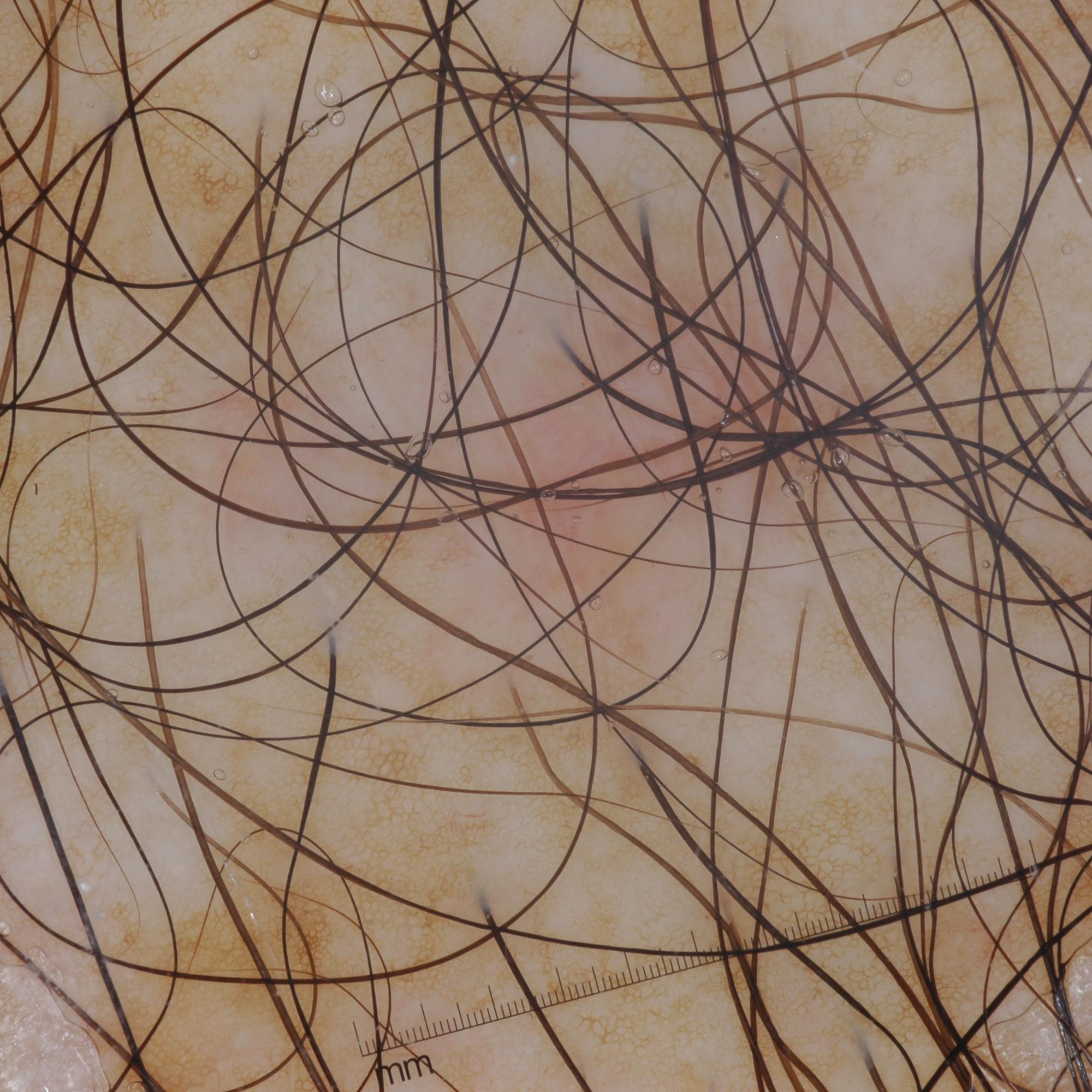}{Tier 5\\(all wrong)}
\caption{One example test image per consensus difficulty tier on H1, where the tier counts how many of the five UQ methods misclassify the sample, from Tier~0 (all five correct) to Tier~5 (all five wrong).}
\label{fig:tier-examples}
\end{figure}

Our main finding is that predictive uncertainty is largely method-agnostic in what it flags. Per-sample entropy is rank-correlated between 0.54 and 0.91 across the methods, so agreement on which lesions are difficult is a property of the samples more than of any method. That agreement has a boundary: on systematically misleading cases, where every model is confidently wrong, entropy dips rather than rises, so uncertainty flags ambiguous cases, not deceptive ones. This is also shown in Figure \ref{fig:tier-examples}.

We make several contributions on uncertainty from a heterogeneous skin lesion dataset: (1) A comparison of five UQ methods (MC Dropout, DropConnect, Flipout, Deep Ensembles, DUQ) against a baseline on a shared backbone, reporting accuracy, calibration (ECE), uncertainty decomposition, and risk-coverage for both heads. (2) Evidence that uncertainty flags the same difficult samples across methods and ranks errors well enough to support selective referral. (3) Evidence that entropy is best predicted by the diagnosis confirmation type, and (4) we show that all methods agree on the indeterminate set being the most uncertain.

We frame the result as support for triage, not autonomous diagnosis. Section~\ref{sec:discussion} details the limits, including a fairness gap on darker skin.

\section{Evaluating Uncertainty as Difficulty Estimation}
\label{sec:methods}

\textbf{An Heterogeneous Evaluation Setting}. We draw all images from the ISIC Archive~\cite{isic_archive,isic_about}, which pools clinical and dermoscopic lesion images from many institutions under a shared metadata format. We keep that mix on purpose. The shared fields merge sources into one dataset with consistent diagnosis labels, while imaging protocols, devices, and patient populations vary by source. A deployed model meets exactly this source-to-source variation, not a single clean distribution.

The pooled set contains $92{,}092$ images from 18 data sources: 17 named institutions plus a large set of anonymous contributions. About $90\%$ are dermoscopic. The rest are clinical close-ups and total-body-photography (TBP) tiles cropped around lesions. As is typical for dermatological data~\cite{cassidyAnalysisISICImage2022}, the classes are imbalanced, with a benign-to-malignant ratio near $70/30$. We hold that ratio fixed across the train, validation, and test splits, so the test set ( 
$5{,}920$ samples) carries the same skew as training.

Fitzpatrick skin type is annotated for only about $15\%$ of images and skews toward lighter types, so we treat it as an analysis variable, never for selection or training. The split guarantees at least $100$ test samples per Fitzpatrick type I--VI before sampling the rest proportionally, keeping every type present even where darker types stay sparse.

\textbf{Task Simplification}. We make two label changes. First, the $3{,}289$ images with an indeterminate primary diagnosis are removed from training and kept as a separate evaluation set, leaving $88{,}803$ benign and malignant images for the train, validation, and test splits. These lesions resist a clean benign-or-malignant call even for clinicians, making them a difficult-sample probe whose difficulty is defined independently of any model. Second, the $22$ fine-grained secondary diagnoses follow a long tail, many with only a handful of images, so we consolidate them into five classes: Benign, Malignant Non-Epidermal, Malignant Epidermal, Melanoma, and Other. The full mapping is in the supplementary material.

\textbf{Multi-task Architecture}. All methods share an ImageNet pretrained \newline EfficientNet-B3 backbone~\cite{tanEfficientNetRethinkingModel2019}, selected by a hyperparameter search over candidate backbones. It feeds two jointly trained heads, each holding the UQ method: a binary head (H1) separating malignant from non-malignant lesions, and a five-class head (H2) over the consolidated categories. Both heads train under a weighted cross-entropy loss summed across heads, with class weights offsetting the imbalance. The joint setup is an implementation choice for this study, not a claimed benefit, since we do not compare it against single-head training.

\textbf{Uncertainty Quantification Methods}. We compare five UQ methods against a plain softmax baseline, spanning the main UQ families: Bayesian approximations, ensembles, and single deterministic methods~\cite{galUncertaintyDeepLearning,gawlikowskiSurveyUncertaintyDeep2023,abdarReviewUncertaintyQuantification2021}. The Baseline is a standard classifier with no UQ method. It makes one deterministic forward pass and serves as a performance and calibration reference. Three methods approximate Bayesian inference by sampling. MC Dropout~\cite{pmlr-v48-gal16} keeps dropout active at test time and averages stochastic passes. DropConnect~\cite{mobinyDropConnectEffectiveModeling2021,wanRegularizationNeuralNetworks2013} masks weights rather than activations. Flipout~\cite{wenFlipoutEfficientPseudoIndependent2018} draws weights from learned distributions using per-example sign perturbations. Deep Ensembles~\cite{lakshminarayananSimpleScalablePredictive} train several independent network copies and read uncertainty off their disagreement. DUQ~\cite{van2020uncertainty} is a single-pass deterministic alternative scoring inputs by distance to learned class centroids in feature space. Each stochastic method uses five forward passes and Deep Ensembles five members, matching the sampling budget. DUQ and the Baseline need one pass.

\textbf{Metrics and Definition of Difficult Sample}. Predictive quality is measured with accuracy on both heads. For calibration we use the expected calibration error (ECE)~\cite{pmlr-v70-guo17a}, the gap between confidence and observed accuracy across bins, and the AUROC of confidence, which measures how well a method's confidence separates correct from incorrect predictions. Total predictive uncertainty is the Shannon entropy of the mean predictive distribution. For the sampling methods (MC Dropout, DropConnect, Flipout, Deep Ensembles) we split it into epistemic and aleatoric parts via the mutual information between predictions and model parameters~\cite{smithUnderstandingMeasuresUncertainty2018}. DUQ is single-pass and gives only a total uncertainty, so it is not decomposed, and the no-UQ Baseline is excluded from this analysis. For selective referral we use risk-coverage analysis~\cite{geifmanSelectiveClassificationDeep2017}: predictions are ranked by entropy, the most uncertain progressively withheld, and error tracked against coverage, summarized by the area under the curve (AURC), lower being better.

We define a difficult sample in two ways. The first is consensus-based: for each input we count how many of the five methods misclassify it, giving a difficulty tier from samples all methods get right to samples all methods miss. Scoring a method against a tier it helped define is circular, so we recompute the tier with a leave-one-model-out (LOMO) scheme: when evaluating a method, it is dropped and difficulty is set by the remaining four. LOMO reduces but does not eliminate the circularity, because the four held-out methods share a backbone and training data. So we lean on a second, model-independent notion: the held-out indeterminate set from Section~\ref{sec:methods}. Those labels came from the data sources themselves, not our models, so higher uncertainty there is a cleaner sign that a method tracks genuine difficulty rather than its own failure modes.

\section{Results}
\label{sec:results}

We report results on the held-out test set ($N\approx5{,}920$) for both heads: H1 (malignant vs.\ non-malignant) and H2 (the five consolidated diagnostic categories). At this sample size every group difference we tested is significant at $p<0.001$, so we read effect sizes rather than $p$-values.

\textbf{Model Comparison}. Table~\ref{tab:comparison} collects accuracy, calibration (ECE), the epistemic share of predictive entropy, confidence AUROC, and risk--coverage area (AURC) for every method on both heads. Four of the five UQ methods match or beat the softmax baseline on accuracy, and Deep Ensembles is the strongest overall: it leads or ties on accuracy and posts the lowest ECE on both heads by a wide margin ($0.013$ on H1, $0.019$ on H2). DropConnect is the opposite case, accurate but the most overconfident of all methods (H2 ECE $0.078$). Flipout is the lone accuracy failure, trailing the baseline by about four points on H1 and more than ten on H2. Its training was unstable, so the low accuracy may reflect the limits of Variational Inference for Bayesian Neural Networks, which is a limitation of this method.

Only Deep Ensembles and MC Dropout turn predictive entropy into a meaningful epistemic component. DropConnect attributes almost none of its entropy to model uncertainty ($\approx3\%$), and Flipout produces near-zero mutual information ($<0.1\%$), so its uncertainty is effectively all aleatoric. DUQ is single-pass with no decomposition. Its entropy comes from normalized RBF-kernel outputs, so the comparison understates the kernel-distance signal DUQ uses for out-of-distribution detection.

\begin{table}[t]
\centering
\caption{Test-set comparison, both heads. ECE = expected calibration error. Epi\% = epistemic share of predictive entropy (mutual information over predictive entropy). AURC = area under the risk--coverage curve. Lower ECE/AURC is better. ``--'' marks no epistemic decomposition (single-pass DUQ and the no-UQ Baseline).}
\label{tab:comparison}
\setlength{\tabcolsep}{4.5pt}
\resizebox{\linewidth}{!}{%
\begin{tabular}{lccccc@{\hspace{1.2em}}ccccc}
\toprule
& \multicolumn{5}{c}{H1 (binary)} & \multicolumn{5}{c}{H2 (multiclass)} \\
\cmidrule(lr){2-6}\cmidrule(lr){7-11}
Method & Acc & ECE & Epi\% & AUROC & AURC & Acc & ECE & Epi\% & AUROC & AURC \\
\midrule
Baseline       & 0.911 & 0.035 & --   & 0.881 & 0.016
               & 0.870 & 0.059 & --   & 0.868 & 0.028 \\
Deep Ensembles & 0.926 & \textbf{0.013} & 16.1 & 0.889 & \textbf{0.012}
               & 0.887 & \textbf{0.019} & 23.7 & 0.884 & 0.022 \\
MC Dropout     & 0.927 & 0.043 & 38.7 & 0.877 & 0.014
               & 0.902 & 0.059 & 43.0 & 0.877 & \textbf{0.020} \\
DropConnect    & 0.927 & 0.057 & 2.6  & 0.885 & 0.014
               & 0.902 & 0.078 & 3.3  & 0.878 & \textbf{0.020} \\
Flipout        & 0.868 & 0.020 & 0.0  & 0.842 & 0.034
               & 0.763 & 0.057 & 0.1  & 0.833 & 0.075 \\
DUQ            & 0.909 & 0.033 & --   & 0.879 & 0.017
               & 0.876 & 0.046 & --   & 0.846 & 0.031 \\
\bottomrule
\end{tabular}%
}
\end{table}

\textbf{Heterogeneity from Pooling Sources}. Pooling many institutional sources brings in clinical heterogeneity (case mix, severity) and technical heterogeneity (imaging protocol, background). A multivariate regression of predictive entropy on metadata ($N=3{,}617$, skin type excluded for low coverage) ranks the drivers (See full details in Tables \ref{tab:mr_entropy_metadata_flipout}, \ref{tab:mr_entropy_metadata_duq_de}, and \ref{tab:mr_entropy_metadata_mcd_mcdc}). Diagnosis confirmation type is the strongest independent predictor of entropy for every method, with effect sizes from about $0.03$ for DropConnect and MC Dropout up to $0.28$ for Flipout. Confirmation type is a workup proxy rather than a direct severity measure, since histopathology cases are biopsied because a clinician judged them suspicious enough to sample. They run 56\% malignant with higher entropy and a stable epistemic share. Entropy also scales with diagnostic difficulty: it is lowest on clearly benign lesions and highest on melanoma and the ambiguous ``Other'' category, the two weakest-accuracy classes, with ``Other'' the lowest of all.

Once confirmation type and diagnosis are controlled for, data source has a moderate independent effect on entropy (partial $\eta^2 \approx 0.01$--$0.03$). The effect is uneven across sources: several show elevated uncertainty, and the Royal Prince Alfred Hospital subset is the clearest outlier, with the highest mean entropy and the lowest accuracy of any source for every method. Its images sit on a distinctive white background where the other sources use black. Deep Ensembles is the most robust there, keeping the highest accuracy ($0.72$ versus $0.63$ for the next-best method). Replacing the white background with black on 16 of these images lowers H1 predictive entropy across all five methods (n=16, too noisy to read accuracy deltas), consistent with a source-specific artifact, though imaging and protocol differences likely also contribute. Per-method deltas are in the supplement.

A second shortcut showed up qualitatively and we did not quantify it. Non-malignant samples with very low entropy almost all carry a clinician-applied measurement marker. The model appears to read the presence of a marker as a cue for non-malignant, which is backwards as a safety property: a marker means a clinician thought the lesion worth measuring, so a confident non-malignant call on a marked lesion is a false-negative risk. We flag this as an observation, not a measured effect.

\begin{figure}[t]
\centering
\crule[cDropConnect]{0.5em}{0.5em} MC-DropConnect \quad \crule[cDropout]{0.5em}{0.5em}  MC-Dropout \quad \crule[cDUQ]{0.5em}{0.5em} DUQ \quad \crule[cEnsemble]{0.5em}{0.5em} Deep Ensemble \quad \crule[cFlipout]{0.5em}{0.5em} Flipout

\begin{tikzpicture}
\begin{groupplot}[
  group style={group size=2 by 1,horizontal sep=1.1cm},
  width=0.47\linewidth,height=4.2cm,
  xlabel={LOMO difficulty tier},xtick={0,1,2,3,4},xmin=-0.2,xmax=4.2,
  tick label style={font=\footnotesize},label style={font=\footnotesize}]
\nextgroupplot[ylabel={Entropy},ymin=0]
\addplot[cEnsemble,mark=*,thick] table[x=tier,y=mean_entropy,col sep=comma]{data/tier_h1_ensemble.csv};
\addplot[cDropout,mark=square*,thick] table[x=tier,y=mean_entropy,col sep=comma]{data/tier_h1_dropout.csv};
\addplot[cDropConnect,mark=triangle*,thick] table[x=tier,y=mean_entropy,col sep=comma]{data/tier_h1_dropconnect.csv};
\addplot[cFlipout,mark=diamond*,thick] table[x=tier,y=mean_entropy,col sep=comma]{data/tier_h1_flipout.csv};
\addplot[cDUQ,mark=x,thick] table[x=tier,y=mean_entropy,col sep=comma]{data/tier_h1_duq.csv};
\nextgroupplot[ylabel={Overconfidence},ymin=0,ymax=1]
\addplot[cEnsemble,mark=*,thick] table[x=tier,y=overconfidence_rate,col sep=comma]{data/tier_h1_ensemble.csv};
\addplot[cDropout,mark=square*,thick] table[x=tier,y=overconfidence_rate,col sep=comma]{data/tier_h1_dropout.csv};
\addplot[cDropConnect,mark=triangle*,thick] table[x=tier,y=overconfidence_rate,col sep=comma]{data/tier_h1_dropconnect.csv};
\addplot[cFlipout,mark=diamond*,thick] table[x=tier,y=overconfidence_rate,col sep=comma]{data/tier_h1_flipout.csv};
\addplot[cDUQ,mark=x,thick] table[x=tier,y=overconfidence_rate,col sep=comma]{data/tier_h1_duq.csv};
\end{groupplot}
\end{tikzpicture}
\vspace{-3pt}
\begin{tikzpicture}
\begin{axis}[hide axis,scale only axis,height=0pt,width=0pt,
  legend columns=5,legend cell align=left,
  legend style={draw=none,font=\footnotesize,
    /tikz/every even column/.append style={column sep=6pt}}]
\addlegendimage{cEnsemble,mark=*,thick}\addlegendentry{Deep Ensembles}
\addlegendimage{cDropout,mark=square*,thick}\addlegendentry{MC Dropout}
\addlegendimage{cDropConnect,mark=triangle*,thick}\addlegendentry{DropConnect}
\addlegendimage{cFlipout,mark=diamond*,thick}\addlegendentry{Flipout}
\addlegendimage{cDUQ,mark=x,thick}\addlegendentry{DUQ}
\end{axis}
\end{tikzpicture}
\caption{Mean predictive entropy (left) and overconfidence (right) by LOMO difficulty tier, H1. The tier-4 entropy dip with peaking overconfidence marks systematically misleading cases (confidently wrong, not uncertain).}
\label{fig:entropy-tier}
\end{figure}
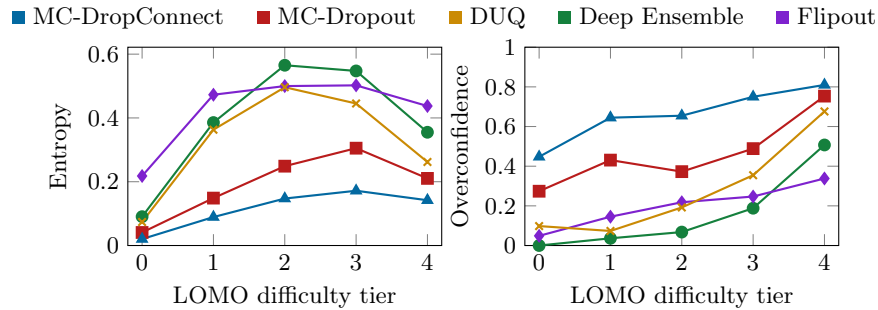

\textbf{Difficult Samples}. Mean predictive entropy rises with the leave-one-model-out (LOMO) difficulty tier for every method, so uncertainty scales with how many models find a sample hard (Fig.~\ref{fig:entropy-tier}). The pattern breaks at the top tier. Entropy dips at tier 4 while overconfidence peaks there: on the samples that nearly all models get wrong, the models are not uncertain but confidently wrong. These are systematically misleading cases, not ambiguous ones. Uncertainty flags the ambiguous hard cases, the ones the model itself recognizes as borderline, and does not flag the cases where the image misleads the model into a confident mistake.

\begin{figure}[t]
\subfloat[Spearman rank correlation,  H1 (lower triangle)\\ and H2 (upper triangle)] {
    \setlength{\tabcolsep}{2pt}
    \begin{tabular}{lccccc}
    \toprule
    & MC-DC & MC-DO & DUQ & Deep Ens. & Flipout. \\
    \midrule
    \crule[cDropConnect]{0.5em}{0.5em} MC-DC    & --    & 0.72  & 0.69  & 0.74  & 0.53 \\
    \crule[cDropout]{0.5em}{0.5em}  MC-DO        & 0.70  & --    & 0.70  & 0.79  & 0.61 \\
    \crule[cDUQ]{0.5em}{0.5em} DUQ            & 0.68  & 0.75  & --    & 0.85  & 0.66 \\
    \crule[cEnsemble]{0.5em}{0.5em} Deep Ens.      & 0.72  & 0.79  & 0.91  & --    & 0.75 \\
    \crule[cFlipout]{0.5em}{0.5em} Flipout        & 0.54  & 0.63  & 0.75  & 0.79  & --   \\
    \bottomrule
    \end{tabular}
    \label{tab:rank-corr}
}
\subfloat[Risk--coverage curves, H1] {
    \centering
    \begin{tikzpicture}
    \begin{axis}[width=0.3\linewidth,height=4cm, xlabel={Coverage},ylabel={Error rate},
  xmin=0,xmax=1,ymin=0,ymax=0.14, tick label style={font=\tiny},label style={font=\tiny},
  legend style={font=\scriptsize,draw=none,fill=none,at={(0.03,0.97)},anchor=north west},
  legend cell align=left, yticklabel style={
        /pgf/number format/fixed,
        /pgf/number format/precision=2
}]
    \addplot[cBaseline,thick,mark=none, cBaseline] table[x=coverage,y=error_rate,col sep=comma]{data/rc_h1_baseline.csv};
    \addplot[cEnsemble,thick,mark=none, cEnsemble] table[x=coverage,y=error_rate,col sep=comma]{data/rc_h1_ensemble.csv};
    \addplot[cDropout,thick,mark=none, cDropout] table[x=coverage,y=error_rate,col sep=comma]{data/rc_h1_dropout.csv};
    \addplot[cDropConnect,thick,mark=none, cDropConnect] table[x=coverage,y=error_rate,col sep=comma]{data/rc_h1_dropconnect.csv};
    \addplot[cFlipout,thick,mark=none, cFlipout] table[x=coverage,y=error_rate,col sep=comma]{data/rc_h1_flipout.csv};
    \addplot[cDUQ,thick,mark=none, cDUQ] table[x=coverage,y=error_rate,col sep=comma]{data/rc_h1_duq.csv};
\end{axis}
\end{tikzpicture}
\label{fig:risk-coverage}
\vspace*{-0.8em}
}
\caption{Left shows spearman rank correlation of per-sample predictive entropy between method pairs, H1 (lower triangle) and H2 (upper triangle). High correlations across very different methods indicate difficulty is mostly carried by the sample. Right shows Risk--coverage curves, H1. Lower error at a given coverage means uncertainty ranks errors better, which selective referral exploits~\cite{mobinyRiskAwareMachineLearning2019,combaliaUncertaintyEstimationDeep2020}. \crule[cBaseline]{0.5em}{0.5em} is a softmax baseline.}
\end{figure}
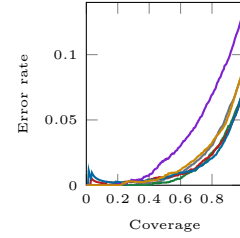

The methods largely agree on which samples are hard. Per-sample Spearman rank correlation of predictive entropy between method pairs ranges from $0.54$ (DropConnect--Flipout) to $0.91$ (DUQ--Deep Ensembles) on H1, with a similar band on H2 (Table~\ref{tab:rank-corr}). Difficulty is largely a property of the sample, not the method: even methods that disagree on absolute entropy level and on calibration still rank the same images as the hardest. The held-out methods used to build LOMO tiers are not independent of the method being scored, exactly because their rankings correlate this strongly.

The circularity-free check comes from the indeterminate set, the 3{,}289 cases held out of training as too ambiguous to label confidently. Every method assigns higher mean entropy to this set than to the test set, and it never informed the tier definitions, so it corroborates the tier result without the LOMO caveat. DropConnect and MC Dropout sit at lower absolute entropy throughout but show the largest relative jump on the indeterminate set.

Ranking by uncertainty supports selective referral. The risk--coverage curves trade coverage for error: deferring the most uncertain predictions to a clinician should remove a disproportionate share of the errors (Fig.~\ref{fig:risk-coverage}). The area under these curves (AURC, lower is better) in Table~\ref{tab:comparison} shows uncertainty ranks errors well enough to be useful, and which method ranks best depends on the head. Deep Ensembles gives the lowest H1 AURC ($0.012$): retaining the $80\%$ most-confident predictions (deferring the most-uncertain $20\%$) lowers the H1 error rate on the retained set from $7.4\%$ at full coverage to $2.5\%$. DropConnect and MC Dropout tie for the lowest on H2 ($0.020$). Flipout is worst on both ($0.034$ on H1, $0.075$ on H2), consistent with its near-absent epistemic signal.

\section{Discussion and Conclusion}
\label{sec:discussion}

\textbf{Difficulty is a property of the sample, not the method}. The clearest result holds across every method we tried. Entropy rises with consensus difficulty tier for all of them, and their per-sample rankings agree closely (Table~\ref{tab:rank-corr}). Methods with narrow entropy distributions, such as DropConnect and MC Dropout, flag the same images as those with wide ones. The method changes how much you can trust a single confidence number, not which samples rank highest by uncertainty. Where the methods diverge is calibration and decomposition. Deep Ensembles has the lowest ECE on both heads by a wide margin and is the only method besides MC Dropout with a non-trivial epistemic share, while DropConnect and Flipout collapse to near-zero epistemic uncertainty (Table~\ref{tab:comparison}). To rank cases by difficulty alone, a much cheaper single-pass method such as DUQ recovers almost the same ordering.

\textbf{Uncertainty would support selective referral, with caveats}. Selective referral needs errors ranked well by uncertainty, and the risk-coverage curves show the ranking is good enough. Deferring the most uncertain cases to a clinician removes a disproportionate share of the model's errors, which drives the low AURC values for Deep Ensembles, DropConnect, and MC Dropout on H1 (See Table~\ref{tab:comparison}). The evidence is in-distribution only: we tested no dedicated out-of-distribution set, so we cannot say how the ranking behaves on inputs unlike the training data. And no clinician reviewed the flagged cases, so we do not know whether they match what an expert finds hard. We therefore read the result as support for an assistive role, flagging uncertain cases for a clinician or acting as a second opinion, not for unsupervised diagnosis.

One boundary condition matters here. Uncertainty flags ambiguous hard cases, but not systematically misleading ones. At the highest difficulty tiers, where most methods are wrong, entropy dips rather than rises and overconfidence peaks: the models are confidently wrong, not uncertain. A policy that defers only high-entropy cases will therefore pass exactly the cases the model fails most consistently. This is a real limit of uncertainty-based triage, not a tuning problem fixed by an operating point.

\textbf{The fairness gap is the central caveat}. The deferral and malignant-detection results hold on a predominantly light-skinned, in-distribution population and cannot extend to darker skin without new data. The raw counts are the reason. The test set has effectively zero malignant samples for Fitzpatrick type VI and only three to six for types IV and V, so any malignant-detection or triage claim for these groups rests on a handful of cases. A type-IV sensitivity computed on six samples is statistically meaningless. One misreading of the entropy data is tempting: darker skin types show lower entropy here, but this tracks the near-absence of malignant cases, not equitable performance. Dermatology models already do worse on darker skin~\cite{daneshjouDisparitiesDermatologyAI2022} and are most accurate on the skin types they were trained on~\cite{grohEvaluatingDeepNeural2021}, which makes it more important to state plainly that we have the least evidence for the populations that need it most.

Across five UQ methods and a baseline on a heterogeneous, multi-source dataset, difficulty is largely method-agnostic, and predictive uncertainty ranks errors well enough to support selective referral as a triage aid. Deep Ensembles performs most consistently when calibration and a meaningful epistemic-aleatoric split matter, and it degrades least on the most affected source. For the difficulty ranking alone, a cheaper single-pass method like DUQ does nearly as well. Several directions would strengthen these claims. A dedicated out-of-distribution evaluation would show whether the ranking survives inputs unlike the training data, where DUQ's kernel distance might beat entropy. Expert review of the flagged cases would test whether the model's uncertainty matches clinical judgement rather than metadata patterns. And balanced malignant data across Fitzpatrick types is the prerequisite for any fairness claim. All results also come from a single EfficientNet-B3 backbone, so they need confirming on other architectures before being generalized.

\section*{Detailed Per-Model Multivariate Regression Results}

This extra section presents full multivariate regression results for Type II ANOVA tests, where $\eta^2$ is the ANOVA effect size, for each UQ method in Table \ref{tab:mr_entropy_metadata_flipout} for Flipout, Table \ref{tab:mr_entropy_metadata_mcd_mcdc} for MC-Dropout and MC-DropConnect, and Table \ref{tab:mr_entropy_metadata_duq_de} for DUQ and Deep Ensembles. Note that for all models, confirmation type is the strongest predictor of entropy.

The model is: $\text{Entropy} = \text{diagnosis}_1 + \text{diagnosis}_2 + \text{diagnosis}_\text{confirm type} + \text{age}_\text{approx} + \text{attribution} + \text{image}_\text{type} + \text{sex} + \text{anatomical site}$ with $N = 3617$.

\begin{table}[]
    \centering
    \caption{Multivariate regression of predictive entropy on metadata variables (Type II SS, Flipout). p-values are denoted as: *: $p < 0.05$, **: $p < 0.01$, ***: $p < 0.001$}
    \resizebox{0.70\linewidth}{!}{
    \begin{tabular}{llllllll}
        \toprule
                  & \multicolumn{3}{c}{Binary Head (H1)} & \multicolumn{3}{c}{Multiclass Head (H2)}\\
         Variable & df & $\eta^2$ & Partial $\eta^2$ & $p$ & $\eta^2$ & Partial $\eta^2$ & $p$ \\
        \midrule
         Confirmation type   & 4 & 0.262 & 0.285 & *** & 0.197 & 0.211 & *** \\
         Secondary diagnosis & 4 & 0.048 & 0.067 & *** & 0.008 & 0.01 & ***\\
         Attribution         & 10 & 0.017 & 0.025 & *** & 0.026 & 0.034 & ***\\
         Image type          & 2 & 0.007 & 0.01 & *** & 0.011 & 0.015 & ***\\
         Anatomical site     & 7 & 0.007 & 0.01 & *** & 0.015 & 0.02 & ***\\
         Age                 & 1 & 0.0 & 0.0 & 0.803 & 0.003 & 0.003 & ***\\
         Primary diagnosis   & 1 & 0.0 & 0.0 & 0.608 & 0.0 & 0.001 & 0.167\\
         Sex                 & 1 & 0.0 & 0.0 & 0.324 & 0.0 & 0.0 & 0.410\\
         \midrule
         Model ($R^2$)       & 30 & & 0.445 & $< 0.001$ &  & 0.35 & $< 0.001$\\
        \bottomrule
    \end{tabular}
    }    
    \label{tab:mr_entropy_metadata_flipout}
\end{table}

\begin{table}[]
    \centering
    \caption{Multivariate regression of predictive entropy on metadata variables (Type II SS) for DUQ and Deep Ensembles). p-values are denoted as: *: $p < 0.05$, **: $p < 0.01$, ***: $p < 0.001$}
    \resizebox{\linewidth}{!}{
    \begin{tabular}{p{1.9cm}lllllllllllll}        
        \toprule
                  & \multicolumn{6}{c}{DUQ} & \multicolumn{6}{c}{Deep Ens.}\\
        \midrule
                  & \multicolumn{3}{c}{H1} & \multicolumn{3}{c}{H2} & \multicolumn{3}{c}{H1} & \multicolumn{3}{c}{H2}\\
         Variable & df & $\eta^2$ & Part. $\eta^2$ & $p$ & $\eta^2$ & Part. $\eta^2$ & $p$ & $\eta^2$ & Part. $\eta^2$ & $p$ & $\eta^2$ & Part. $\eta^2$ & $p$ \\
        \midrule
         Confirmation type   & 4 & 0.069 & 0.073 & *** & 0.103 & 0.106 & *** & 0.143 & 0.152 & *** & 0.176 & 0.188 & *** \\
         Secondary diagnosis & 4 & 0.021 & 0.024 & *** & 0.003 & 0.003 & 0.033 & 0.027 & 0.033 & *** & 0.006 & 0.008 & ***\\
         Attribution         & 10 & 0.015 & 0.017 & *** & 0.012 & 0.014 & *** & 0.021 & 0.025 & *** & 0.023 & 0.029 & ***\\
         Image type          & 2 & 0.006 & 0.007 & *** & 0.005 & 0.005 & *** & 0.008 & 0.01 & *** & 0.013 & 0.017 & ***\\
         Anatomical site     & 7 & 0.008 & 0.009 & *** & 0.008 & 0.009 & *** & 0.006 & 0.007 & *** & 0.015 & 0.019 & ***\\
         Age                 & 1 & 0.004 & 0.004 & *** & 0.004 & 0.005 & *** & 0.002 & 0.002 & 0.004 & 0.004 & 0.006 & ***\\
         Primary\newline diagnosis   & 1 & 0.0 & 0.0 & 0.494 & 0.001 & 0.001 & 0.103 & 0.0 & 0.0 & 0.201 & 0.001 & 0.001 & .107\\
         Sex                 & 1 & 0.0 & 0.0 & 0.863 & 0 & 0 & 0.641 & 0 & 0 & 0.307 & 0.0 & 0.0 & 0.308\\
         \midrule
         Model ($R^2$)       & 30 & & 0.211 & *** &  & 0.183 & *** & & 0.291 & *** & & 0.308 & ***\\
        \bottomrule
    \end{tabular}
    }    
    \label{tab:mr_entropy_metadata_duq_de}
\end{table}

\begin{table}[]
    \centering
    \caption{Multivariate regression of predictive entropy on metadata variables (Type II SS) for MC-Dropout and MC-DropConnect). p-values are denoted as: *: $p < 0.05$, **: $p < 0.01$, ***: $p < 0.001$}
    \resizebox{\linewidth}{!}{
    \begin{tabular}{p{1.9cm}lllllllllllll}        
        \toprule
                  & \multicolumn{6}{c}{MC-Dropout} & \multicolumn{6}{c}{MC-DropConnect}\\
        \midrule
                  & \multicolumn{3}{c}{H1} & \multicolumn{3}{c}{H2} & \multicolumn{3}{c}{H1} & \multicolumn{3}{c}{H2}\\
         Variable & df & $\eta^2$ & Part. $\eta^2$ & $p$ & $\eta^2$ & Part. $\eta^2$ & $p$ & $\eta^2$ & Part. $\eta^2$ & $p$ & $\eta^2$ & Part. $\eta^2$ & $p$ \\
        \midrule
         Confirmation type   & 4 & 0.025 & 0.026  & *** & 0.031 & 0.033 & *** & 0.025 & 0.026 & *** & 0.028 & 0.029 & *** \\
         Secondary diagnosis & 4 & 0.004 & 0.04 & 0.003 & 0.011 & 0.012 & *** & 0.004 & 0.005 & 0.003 & 0.005 & 0.006 & ***\\
         Attribution         & 10 & 0.013 & 0.013 & *** & 09.01 & 0.011 & *** & 0.024 & 0.025 & *** & 0.017 & 0.018 & ***\\
         Image type          & 2 & 0.003 & 0.003 & 0.003 & 0.005 & 0.005 & *** & 0.012 & 0.013 & *** & 0.01 & 0.01 & ***\\
         Anatomical site     & 7 & 0.006 & 0.007 & 0.001 & 0.011 & 0.011 & *** & 0.004 & 0.028 & 0.006 & 0.006 & 0.003\\
         Age                 & 1 & 0.002 & 0.002 & 0.004 & 0.006 & 0.007 & *** & 0.003 & 0.003 & 0.002 & 0.006 & 0.006 & ***\\
         Primary\newline diagnosis   & 1 & 0.0 & 0.0 & 0.405 & 0.001 & 0.001 & 0.093 & 0.0 & 0.0 & 0.378 & 0 & 0 & 0.42\\
         Sex                 & 1 & 0.0 & 0.0 & 0.024 & 0.001 & 0.001 & 0.127 & 0.0 & 0.0 & 0.305 & 0.001 & 0.001 & 0.119\\
         \midrule
         Model ($R^2$)       & 30 & & 0.099 & *** &  & 0.132 & *** & & 0.084 & *** & & 0.088 & ***\\
        \bottomrule
    \end{tabular}
    }    
    \label{tab:mr_entropy_metadata_mcd_mcdc}
\end{table}

\section*{Entropy Growth in Indeterminate Test Set}
This section presents numerical results on how entropy behaves on the indeterminate test set. We expect that the indeterminate set produces higher entropy, showing the difficulty of recognizing this unlabeled class. Results are presented in Table \ref{tab:entropy_indet}, on all UQ methods we see a significant entropy increase in the indeterminate test set.
\begin{table}[!h]
    \centering
    \caption{Mean predictive entropy for test and indeterminate sets, with relative increase ($\Delta \%$). A larger $\Delta \%$ means the method responds more to sample difficulty.}
    \resizebox{0.75\linewidth}{!}{
    \begin{tabular}{lllrp{0.3cm}llr}        
        \toprule
                  & \multicolumn{3}{c}{Binary (H1)}  & & \multicolumn{3}{c}{Multiclass (H2)}\\
        \midrule
         UQ Method      & Test Set & Indet. Set & $\Delta \%$ & & Test Set & Indet. Set & $\Delta \%$ \\
        \midrule
         Deep Ensembles & 0.160 & 0.392 & 145.9\% & & 0.236 & 0.651 & 176.3\%\\
         MC-Dropout     & 0.071 & 0.233 & 228.3\% & & 0.096 & 0.388 & 253.4\%\\
         MC-DropConnect & 0.039 & 0.144 & 268.5\% & & 0.050 & 0.202 & 305.6\%\\
         Flipout        & 0.255 & 0.420 & 64.2\% & & 0.457 & 0.838 & 83.6\%\\
         DUQ            & 0.134 & 0.337 & 151.8\% & & 0.204 & 0.497 & 144.2\%\\
        \bottomrule
    \end{tabular}
    }    
    \label{tab:entropy_indet}
\end{table}
\begin{table}[!hb]
    \centering
    \caption{Reproduction of Table \ref{tab:rank-corr} with 95\% confidence intervals using the Fisher approximation. H1 (lower triangle) and H2 (upper triangle).}
    \resizebox{0.8\linewidth}{!}{    
    \begin{tabular}{llllll}       
        \toprule
        & MC-DC & MC-DO & DUQ & Deep Ens. & Flipout. \\
        \midrule
        MC-DC      & --    & 0.72 [0.71, 0.73]  & 0.69 [0.68, 0.70]  & 0.74 [0.73, 0.75]  & 0.53 [0.52, 0.55] \\
        MC-DO      & 0.70 [0.69, 0.71]  & --    & 0.70 [0.69, 0.71]  & 0.79 [0.78, 0.80]  & 0.61 [0.59, 0.62] \\
        DUQ        & 0.68 [0.67, 0.69]  & 0.75 [0.74, 0.76]  & --    & 0.85 [0.85, 0.86]  & 0.66 [0.64, 0.67] \\
        Deep Ens.  & 0.72 [0.71, 0.74]  & 0.79 [0.78, 0.80]  & 0.91 [0.90, 0.91]  & --    & 0.75 [0.74, 0.76] \\
        Flipout    & 0.54 [0.52, 0.56]  & 0.63 [0.61, 0.65]  & 0.75 [0.74, 0.76]  & 0.79 [0.78, 0.80]  & --   \\
        \bottomrule
    \end{tabular}
    }    
\end{table}

\clearpage
\bibliographystyle{splncs04}
\bibliography{references}

\clearpage

\section*{Supplementary Material}

\subsection*{S1. Background-swap experiment}

A recurring concern with the Royal Prince Alfred Hospital subset is that its uniform white background acts as a shortcut. To probe this, we took the 16 images from that source and replaced the white background with black, then repeated the experiment with a second variant in which the surgical skin markers were also painted out. Table~\ref{tab:bgswap} reports the mean predictive entropy on the original images and the change ($\Delta$) after each edit, for both the binary head (H1) and the five-class head (H2).

\begin{table}[ht]
\centering
\caption{Mean predictive entropy on the 16 Prince Alfred images and its change after editing the background. Lower entropy means the model became more confident. ``Black bg'' replaces the white background with black. ``+markers removed'' additionally erases the surgical markers.}
\label{tab:bgswap}
\begin{tabular}{l cc cc cc}
\toprule
& \multicolumn{2}{c}{Original} & \multicolumn{2}{c}{Black bg ($\Delta$)} & \multicolumn{2}{c}{+markers removed ($\Delta$)} \\
\cmidrule(lr){2-3}\cmidrule(lr){4-5}\cmidrule(lr){6-7}
Method & H1 & H2 & H1 & H2 & H1 & H2 \\
\midrule
Deep Ensembles & 0.504 & 0.475 & $-0.074$ & $+0.068$ & $-0.114$ & $+0.075$ \\
MC Dropout     & 0.224 & 0.234 & $-0.066$ & $-0.045$ & $-0.097$ & $-0.099$ \\
DropConnect    & 0.283 & 0.286 & $-0.219$ & $-0.188$ & $-0.158$ & $-0.154$ \\
Flipout        & 0.590 & 0.805 & $-0.119$ & $-0.106$ & $-0.135$ & $-0.099$ \\
DUQ            & 0.408 & 0.462 & $-0.092$ & $-0.052$ & $-0.174$ & $-0.052$ \\
\bottomrule
\end{tabular}
\end{table}

With only $n=16$ images the result is suggestive rather than conclusive. Entropy on H1 falls for all five methods once the white background is gone, and H2 entropy falls for four of the five. Deep Ensembles is the exception: its H2 entropy rises slightly in both variants. Removing the markers on top of the background swap deepens the drop for most methods but does not change the direction. We do not report accuracy deltas, because at $n=16$ each flipped case moves accuracy by about 6.3 points, so the numbers are too noisy to interpret. The pattern fits a source-specific background artifact, yet because the edit also changes lighting and contrast around the lesion, it is not a clean isolation of the background alone.

\subsection*{S2. Consolidated diagnosis mapping}

The dataset records 22 fine-grained secondary diagnosis categories, which we collapse into five classes for the H2 task. The groupings are:

\noindent\textbf{Benign.} Benign melanocytic proliferations, benign epidermal proliferations, benign (other), benign soft tissue proliferations (fibro-histiocytic, vascular, neural), benign adnexal epithelial proliferations (sebaceous, follicular, apocrine or eccrine), cysts, hemorrhagic lesions, and mast cell proliferations.

\noindent\textbf{Melanoma.} Malignant melanocytic proliferations (melanoma).

\noindent\textbf{Malignant Epidermal.} Malignant epidermal proliferations.

\noindent\textbf{Malignant Non-Epidermal.} Malignant adnexal epithelial proliferations (follicular, apocrine or eccrine) and malignant soft tissue proliferations (fibro-histiocytic, vascular).

\noindent\textbf{Other.} Indeterminate epidermal and melanocytic proliferations, flat melanotic pigmentations that are not melanocytic nevi, inflammatory or infectious diseases, collisions (benign-only, or with at least one malignant component), and exogenous.

\subsection*{S3. Qualitative examples}

\begin{figure}[h]
\noindent\textbf{Benign}\par
\noindent%
\includegraphics[width=0.14\linewidth]{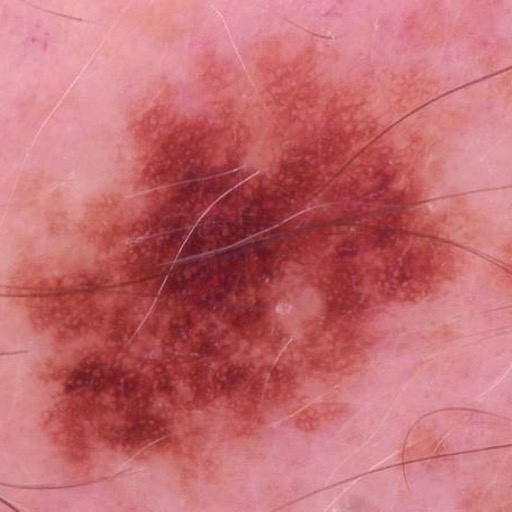}\hfill
\includegraphics[width=0.14\linewidth]{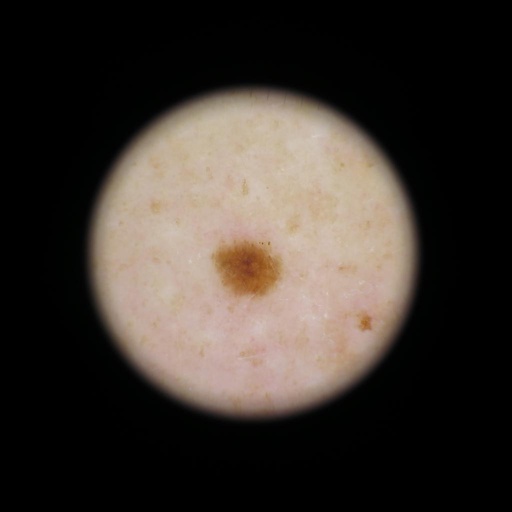}\hfill
\includegraphics[width=0.14\linewidth]{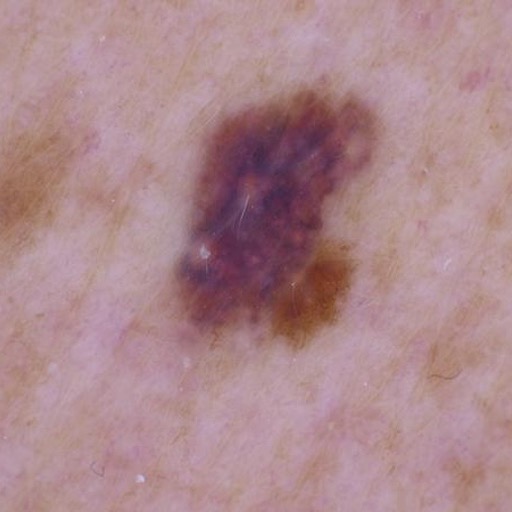}\hfill
\includegraphics[width=0.14\linewidth]{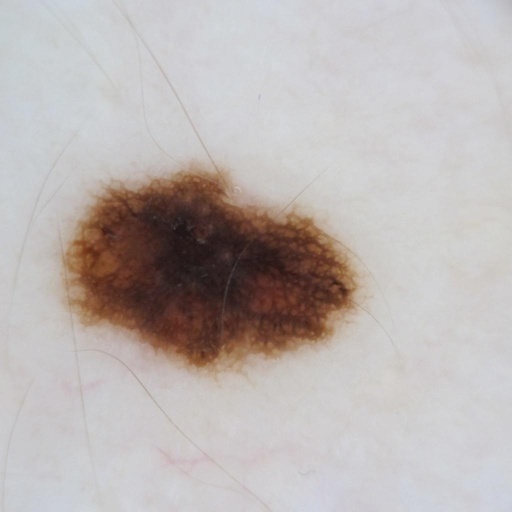}\hfill
\includegraphics[width=0.14\linewidth]{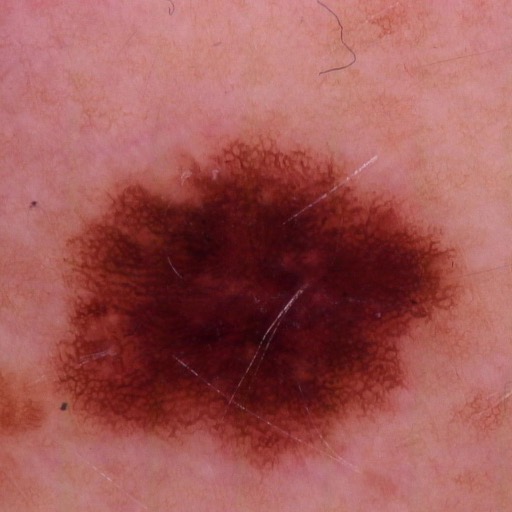}\hfill
\includegraphics[width=0.14\linewidth]{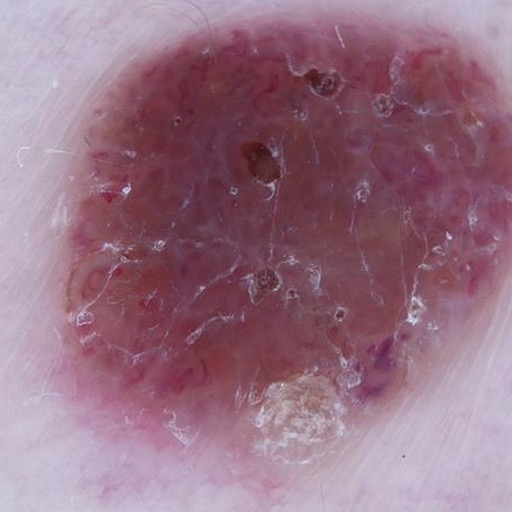}\par
\noindent\textbf{Malignant}\par
\noindent%
\includegraphics[width=0.14\linewidth]{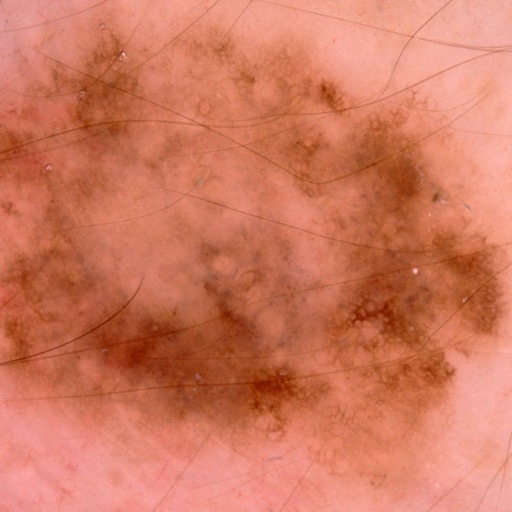}\hfill
\includegraphics[width=0.14\linewidth]{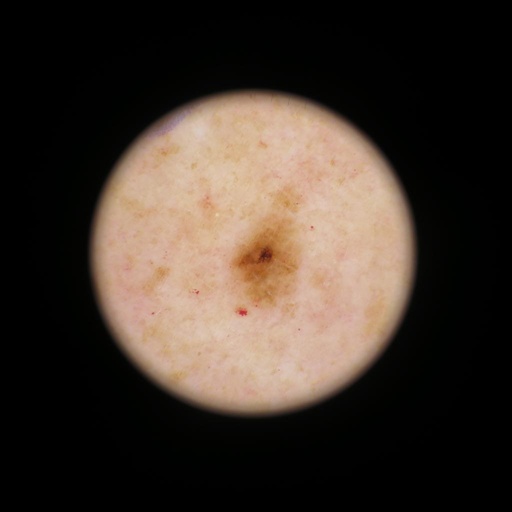}\hfill
\includegraphics[width=0.14\linewidth]{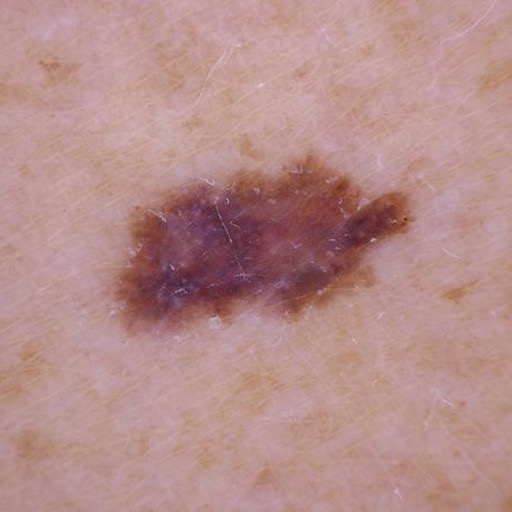}\hfill
\includegraphics[width=0.14\linewidth]{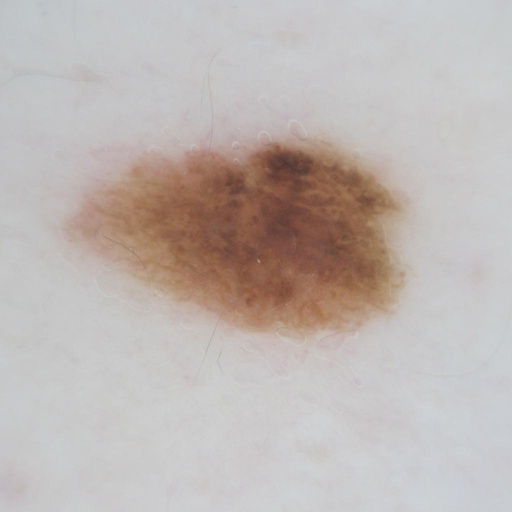}\hfill
\includegraphics[width=0.14\linewidth]{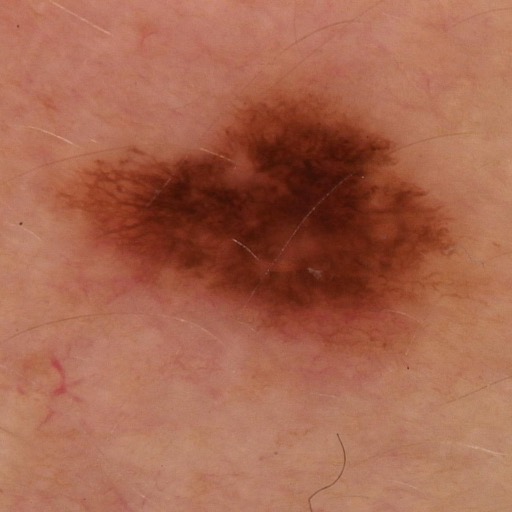}\hfill
\includegraphics[width=0.14\linewidth]{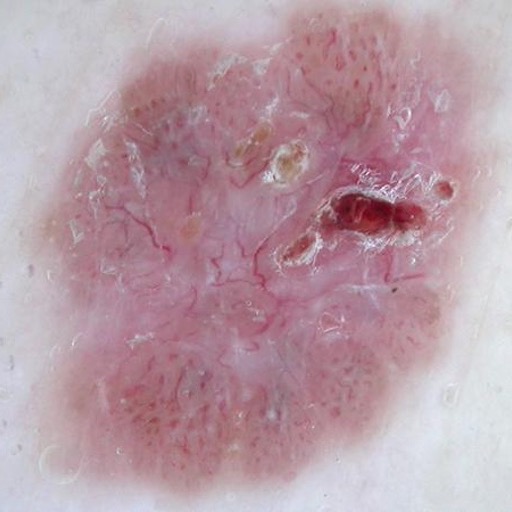}
\caption{Examples of visually similar benign--malignant lesion pairs from the ISIC archive. Each column shows a benign (top) and malignant (bottom) lesion paired by similarity in a neural network's learned feature representation. The visual diversity across columns demonstrates intra-class variation.}
\label{fig:interclass-similarity}
\end{figure}

\begin{figure}[h]
\centering
\newcommand{\dcell}[2]{%
  \begin{minipage}[t]{0.152\linewidth}\centering
  \includegraphics[width=\linewidth]{#1}\\[2pt]
  {\scriptsize #2}\end{minipage}}
\dcell{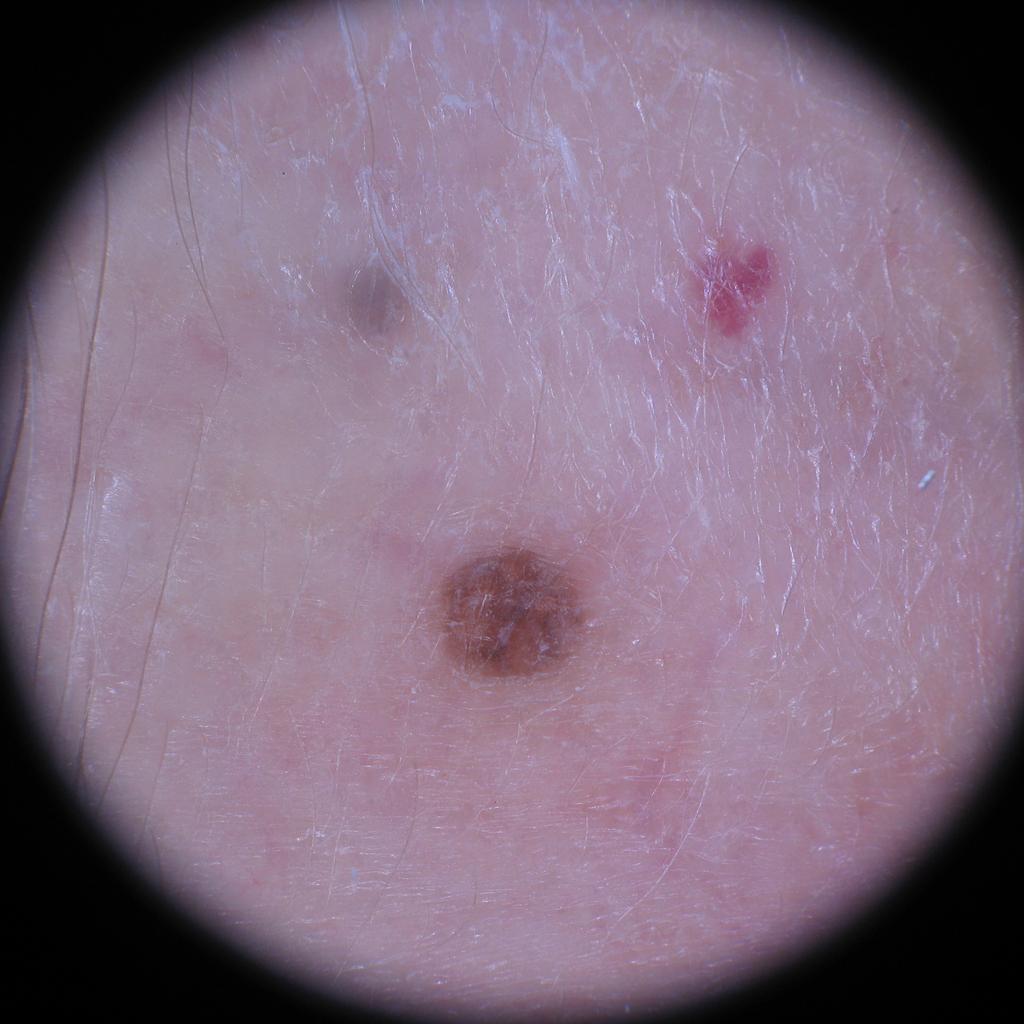}{Hosp.\ Cl\'inic Barcelona}\hfill
\dcell{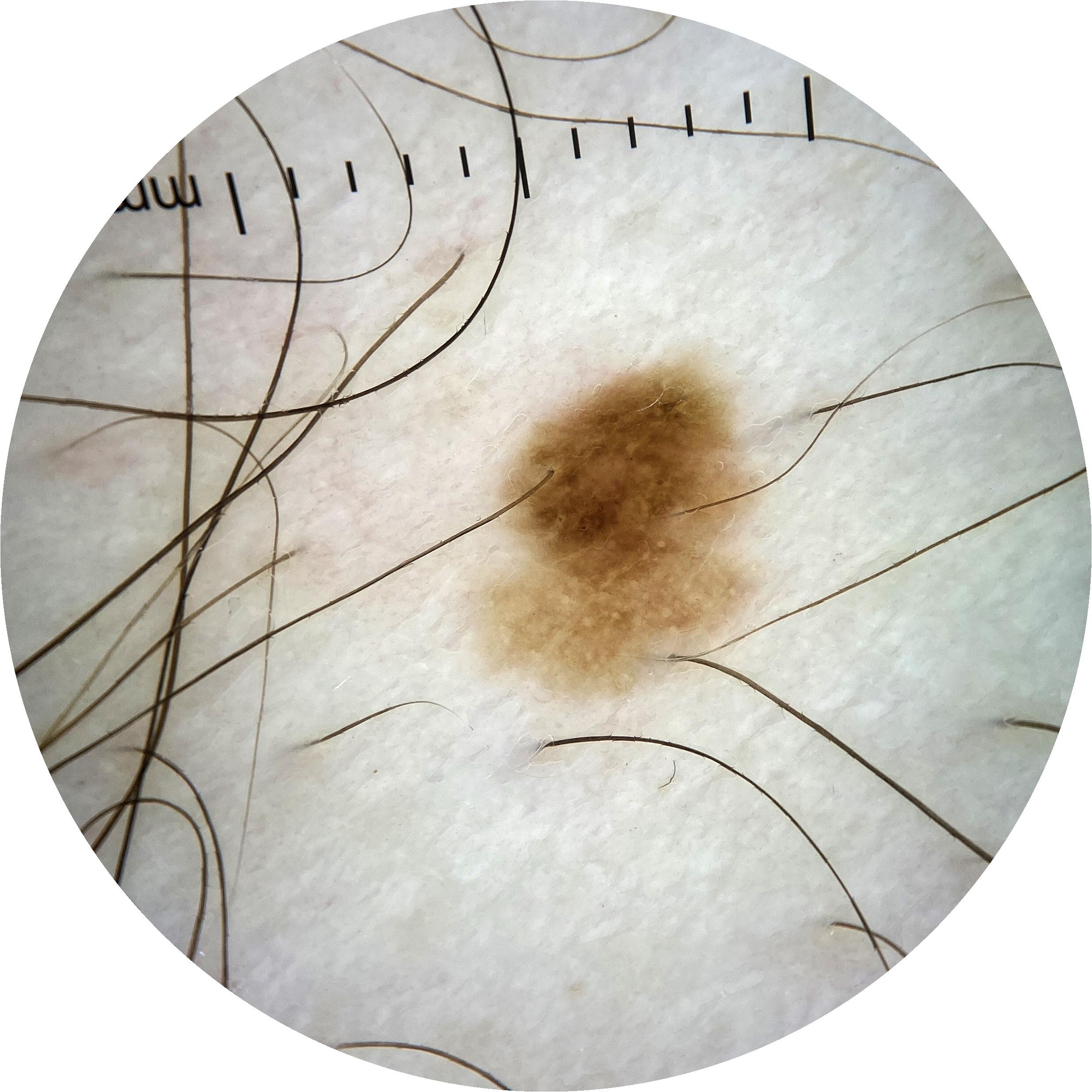}{Royal Prince Alfred Hosp.}\hfill
\dcell{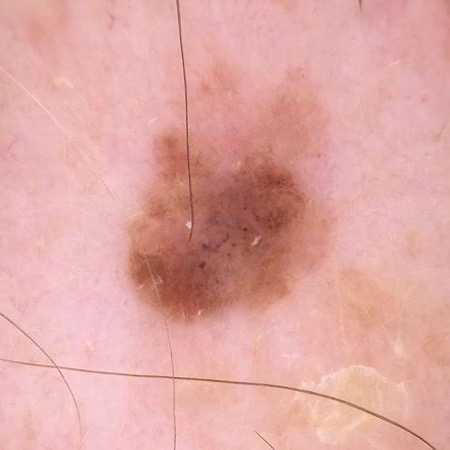}{MILK Study}\hfill
\dcell{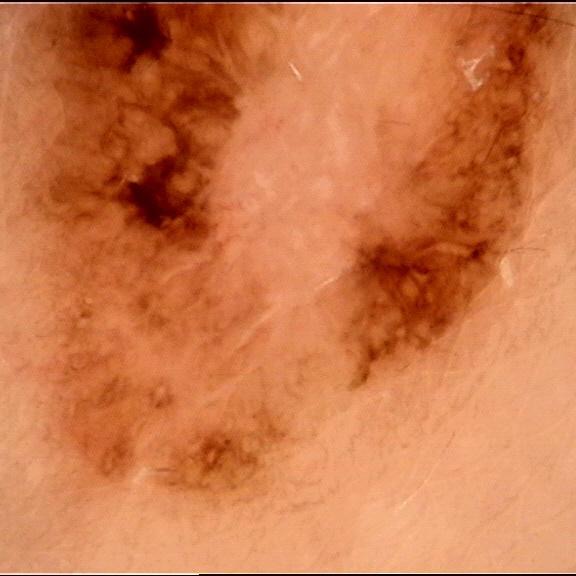}{Imperial College London}\hfill
\dcell{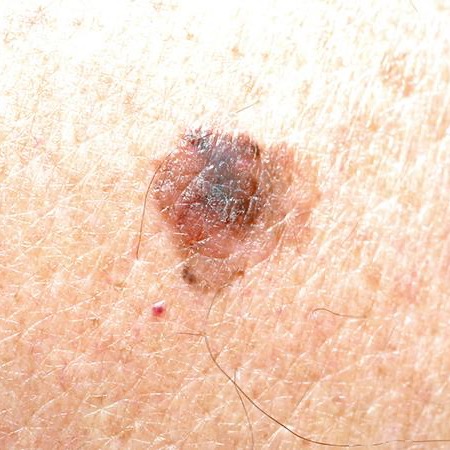}{MILK Study (clinical)}\hfill
\dcell{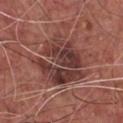}{Memorial Sloan Kettering}
\caption{Six melanoma images from different data sources and imaging modalities. The images differ substantially in background color, magnification, lighting, and resolution due to differences in clinical equipment and imaging protocols.}
\label{fig:dist-shift}
\end{figure}

\begin{figure}[hb]
\centering
\includegraphics[width=0.15\linewidth]{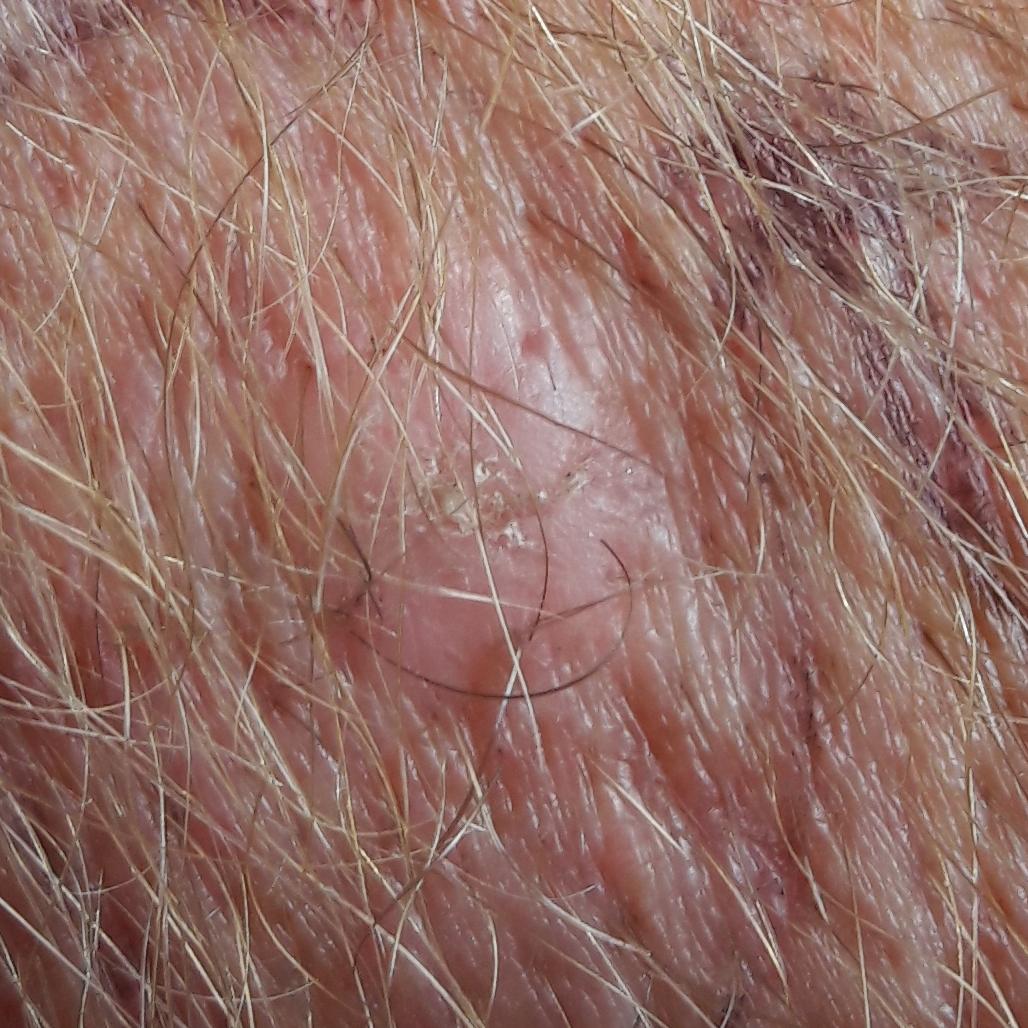}\hfill
\includegraphics[width=0.15\linewidth]{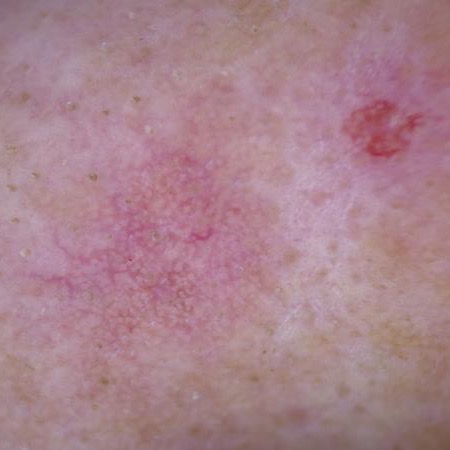}\hfill
\includegraphics[width=0.15\linewidth]{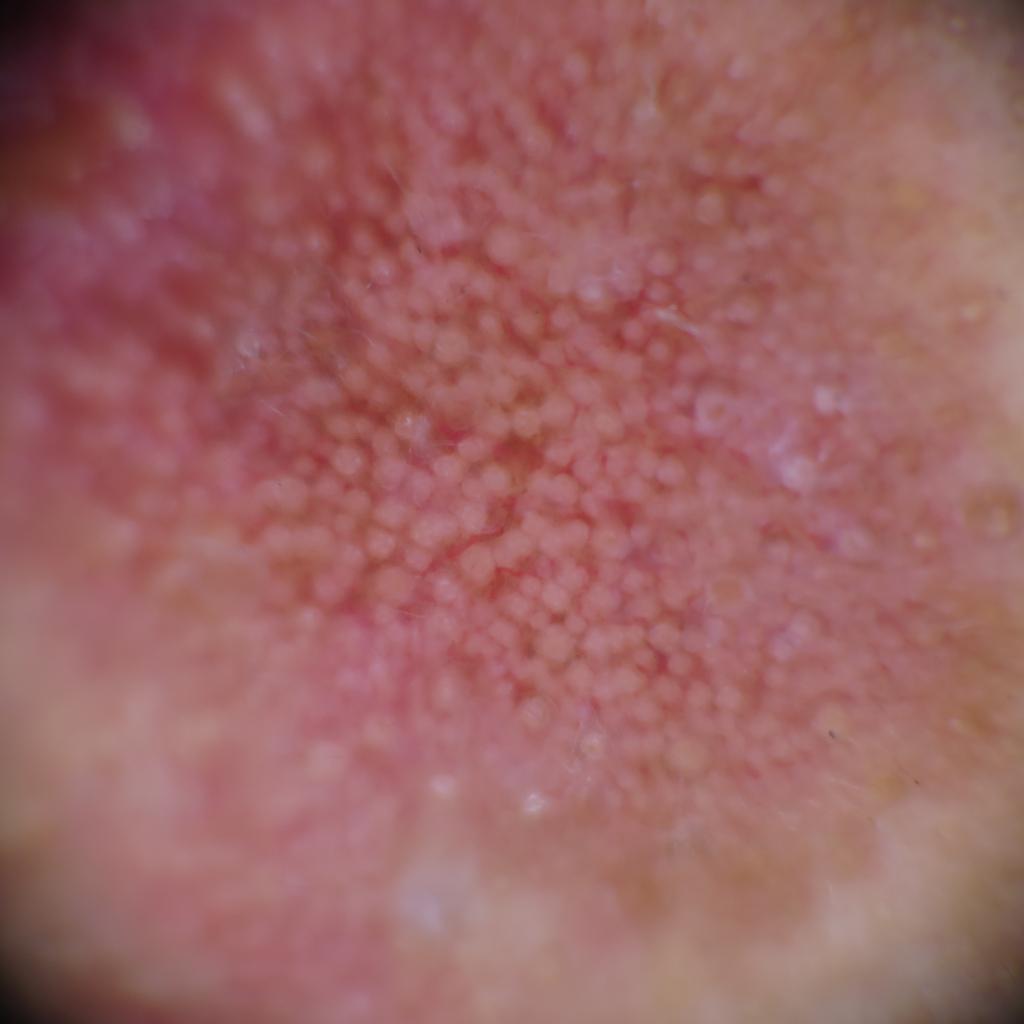}\hfill
\includegraphics[width=0.15\linewidth]{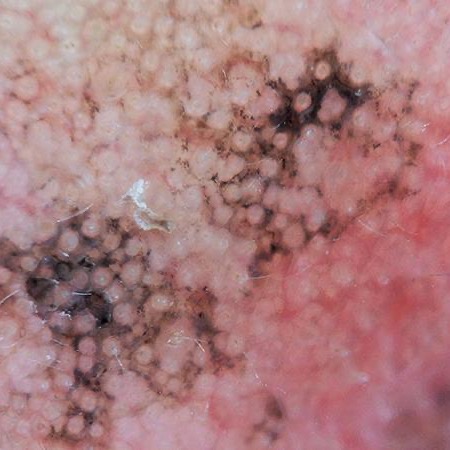}\hfill
\includegraphics[width=0.15\linewidth]{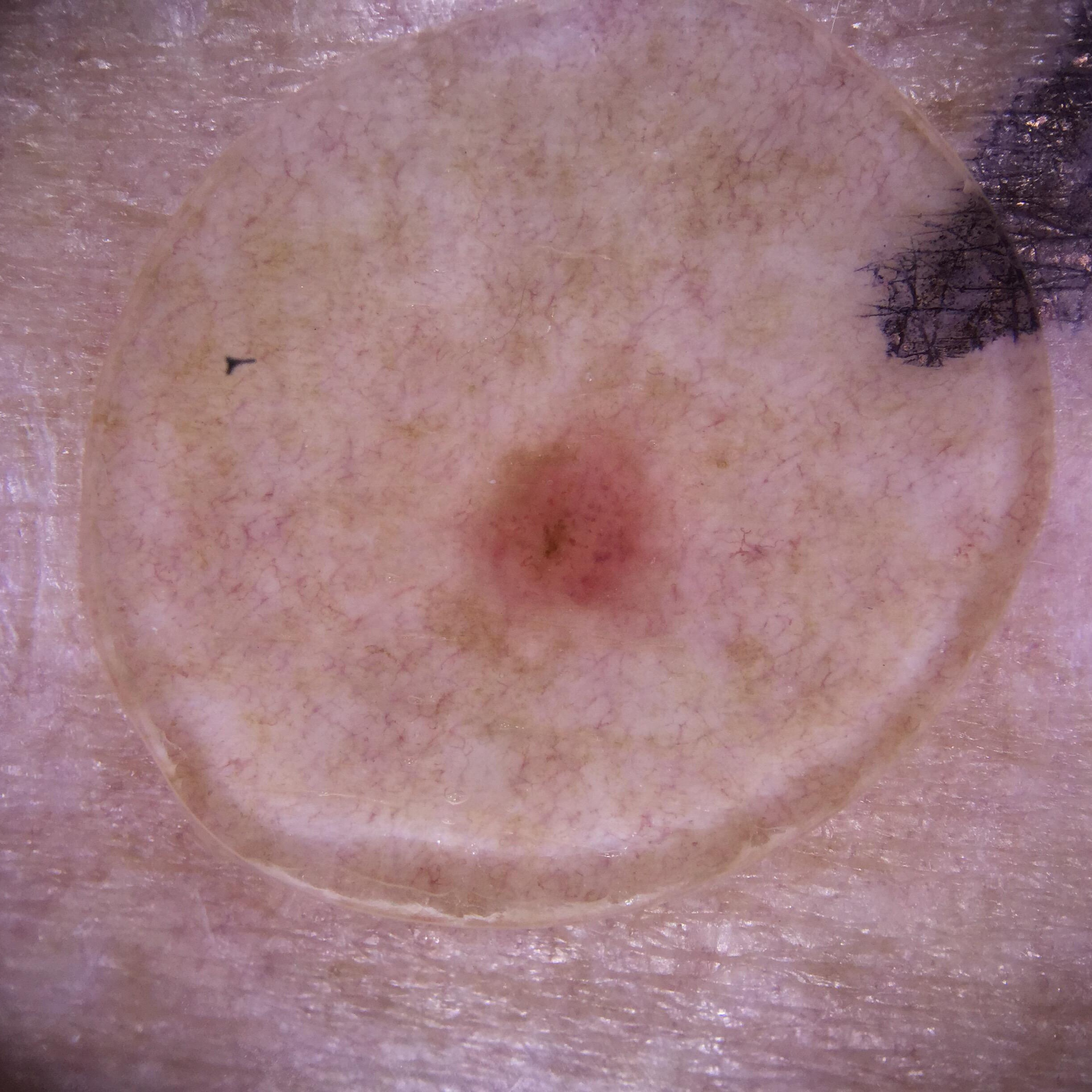}
\caption{Five indeterminate-set samples with the highest predictive entropy on the multiclass head, ranked by Deep Ensembles.}
\label{fig:indet-entropy}
\end{figure}

\end{document}